\pdfoutput=1

\documentclass[11pt]{article}

\usepackage[review]{naacl2021}

\usepackage{times}
\usepackage{latexsym}
\usepackage{mathtools}

\usepackage[T1]{fontenc}

\usepackage[utf8]{inputenc}

\usepackage{microtype}

\usepackage{graphicx}
\usepackage{booktabs}

\title{Leveraging Deep Representations of Radiology Reports in Survival Analysis for Predicting Heart Failure Patient Mortality}

\author{
Hyun Gi Lee$^1$\hspace{1em} Evan Sholle$^2$\hspace{1em} Ashley Beecy$^3$\hspace{1em} Subhi Al'Aref$^4$\hspace{1em} Yifan Peng$^1$\\
$^1$Department of Population Health Sciences, Weill Cornell Medicine \\
$^2$Information Technologies and Services, Weill Cornell Medicine \\
$^3$Department of Medicine, Weill Cornell Medicine \\
$^4$Department of Internal Medicine, UAMS College of Medicine \\
}

\begin{document}
\maketitle
\begin{abstract}
Utilizing clinical texts in survival analysis is difficult because they are largely unstructured. Current automatic extraction models fail to capture textual information comprehensively since their labels are limited in scope. Furthermore, they typically require a large amount of data and high-quality expert annotations for training. In this work, we present a novel method of using BERT-based hidden layer representations of clinical texts as covariates for  proportional hazards models to predict patient survival outcomes. We show that hidden layers yield notably more accurate predictions than predefined features, outperforming the previous baseline model by 5.7\% on average across C-index and time-dependent AUC. We make our work publicly available at \url{https://github.com/bionlplab/heart_failure_mortality}.
\end{abstract}
\section{Introduction}

Survival analysis estimates the expected time until an event of interest occurs~\cite{ranganath2016deep}. In clinical research, it is used to understand the relationship between prognostic covariates (e.g., age and treatment) and patient survival time for important use cases such as predicting mortality of heart failure patients and providing management recommendations for intensive care units during a public health crisis like the COVID-19 pandemic \cite{pandey,sprung2020adult,nielsen2019survival}.

Clinical texts such as radiology reports contain rich information about patients that is used to diagnose disease, plan treatments and monitor progress. It also contains the high-level reasoning of human experts that requires years of knowledge accumulation and professional training \cite{langlotz2015radiology}. Despite their clinical relevance, it is challenging to use them in survival analysis since they are largely unstructured. Automatic labelers are often unable to capture detailed information to distinguish between patients, especially ones with similar conditions, because they mostly rely on a small set of manually selected labels \cite{lao2017deepa}. Developing methods for accessing the critical information embedded in unstructured clinical texts holds the potential to meaningfully benefit clinical research.

To bridge this gap, we propose a deep learning method to predict the survival probability of heart failure (HF) patients based on the high-dimensional feature representations of their radiology reports. Concretely, we extract hidden features from the texts with BERT-based \citep{devlin-etal-2019-bert} models and apply a recurrent neural network (RNN) to model sequences of reports and estimate the log-risk function for the overall mortality prediction. This approach can encapsulate more textual information than hand-crafted features and incorporate higher-order temporal information from report sequences. We find that our model improves on average 5.7\% in both C-index and time-dependent AUC without requiring additional expert annotations. 

We make three contributions through this work: (1) present a novel survival analysis model to leverage feature representations from clinical texts, (2) demonstrate that our model outperforms the ones dependent on predefined expert features and that this approach can generalize across various biomedical and clinical BERT models, and (3) make our work publicly available for reproduction by others. 

\section{Related Work}
Due to the lack of expert annotations, earlier automatic labelers mostly use predefined linguistic patterns to extract relevant information. NegEx \citep{chapman:2001} is a regular expression algorithm that identifies observations based on specified phrases. NegBio \citep{negbio} uses universal dependencies and subgraph matching in addition to regular expressions. CheXpert \cite{irvin:2019} extends NegBio by adding rules to extract, classify, and aggregate mentions to improve performance. While they typically achieve a high precision, they suffer from a low recall because of their limited rules.

BERT \citep{devlin-etal-2019-bert} is a transformer-based method that extracts feature representations of unlabeled text that are effective for transfer learning across various NLP tasks. It is adapted to a wide range of domains, including biomedical and clinical domains \citep{biobert,clinicalbert, bluebert}. Recently, BERT models have been applied to labeling radiology reports. CheXbert \citep{smit:2020} and CheXpert++~\citep{chexpert_plus} train on silver-standard datasets created with a rule-based labeler, CheXpert. Although they outperform rule-based labelers, these approaches need a curated training corpus which can be costly to obtain and error-prone. Furthermore, their labels are still limited and can miss critical information from the reports.

Regarding survival analysis, the Cox proportional hazards model (CPH) is widely adopted as it can deal with censored data and evaluate the prognostic values of covariates simultaneously \citep{cox}. DeepSurv \citep{Katzman_2018} and DeepHit \citep{lee2018deephit} are more contemporary methods that use deep neural networks to model more complex, nonlinear relationships of predictor variables. RNN-SURV \citep{Giunchiglia2018RNNSURVAD} and DRSA \citep{ren2019deep} model time-variant, sequential patterns from predictor variables. To the best of our knowledge, the compatibility of these models and high-dimensional features as covariates has not been tested. 

Automatic extraction tools enable survival analysis to incorporate textual information from clinical texts. \citet{pandey} used a convolutional neural network to extract findings from radiology reports of heart failure patients and predict all-cause mortality with CPH. \citet{heo-etal-2020-various} performed stroke prognosis based on the document-level and sentence-level representations of MRI records. Our work extends this line of research by using contextual deep representations of clinical texts to perform survival analysis.

\section{Methods}

\subsection{Task}

We first formulate the survival analysis problem. In the discrete context, we divide the continuous time into disjoint intervals  $V = (t_{l-1}, t_l]$ where $t_0$ and $t_T$ are the first and last observation interval boundaries. At time $t_u$, the model predicts the survival probability in the prediction window $(t_u,t_T]$ with longitudinal features in the observation window $(t_0,t_u]$ (Figure~\ref{fig:architecture}). 

For each participant $i$, the survival probability at each time $t_l$ ($l>u$) is $S_i(t_l) = Pr(z>t_l)$, where $z$ is the time-to-event, time until death in our case. The hazard rate of the survival probability is $\lambda_i(t_l) = \frac{S_i(t_{l-1})-S_i(t_l)}{S_i(t_l)}$. 

\subsection{Model}

\begin{figure}[!hbpt]
  \vspace{-.5em}
  \includegraphics[width=\linewidth,page=2,clip,trim=1em 2em 10em 9em]{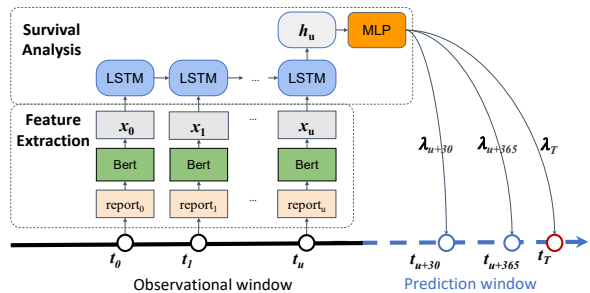}
  \vspace{-1em}
  \caption{Model Architecture.}
  \label{fig:architecture}
  \vspace{-1em}
\end{figure}

Our framework consists of two stages: feature extraction and survival analysis (Figure~\ref{fig:architecture}).

\subsubsection{Feature Extraction}

The input of each time $t_l$ is given by the features extracted from the reports of each patient $i$. In this work, we evaluate two sets of predefined features and hidden features of the reports. 

The first feature set consists of 14 common radiographic findings in computed tomography (CT) imaging reports (aortic aneurysm, ascites, atelectasis, atherosclerosis, cardiomegaly, enlarged liver, gall bladder wall thickening, hernia, hydronephrosis, lymphadenopathy,  pleural effusion, pneumonia, previous surgery, and pulmonary edema). The findings are extracted using the convolutional neural network provided by \citet{pandey} which had the reported performance of 0.90 F1 in average. 

The second feature set consists of 14 predefined findings in CheXpert \citep{irvin:2019} which are commonly found in radiology reports (atelectasis, cardiomegaly, consolidation, edema, enlarged cardiomediastinum, fracture, lung lesion, lung opacity, pleural effusion, pleural other, pneumonia, pneumothorax, support devices, and normal). These features are extracted using CheXbert \citep{smit:2020} with the reported performance of 0.80 F1 in average. 

For deep representations, we use CheXbert’s final hidden layer features of the reports. More specifically, we extract the information before it passes on to the output layer that consists of 14 linear heads. The representations are vectors of size 768. 

Lastly, we construct sequential deep representations by creating arrays of up to three most recent reports of each patient. As the reports can change over time based on the patient's condition, these features are time-variant and contain temporal information that cannot be obtained by a single report. In addition to CheXbert, we apply BERT variations -- BERT, BioBert, ClinicalBert and BlueBert.

\subsubsection{Deep survival analysis}

In this study, the hazard rate has the form
\begin{equation}
    \lambda_i(t_l \mid X_i) = \lambda_u(t_l) e^{\psi(X_i)}
\end{equation}
$\psi$ is a patient's log-risk of failure, $X_i$ are covariates representing a patient’s variables up to $t_u$, and $\lambda_u$ baseline hazard at $t_u$. 

For the standard Cox Proportional-Hazards (CPH) model \citep{cox}, $\psi(X_i)$ has the form of a linear combination of $p$ covariates $\beta_1 X_{i1} + \cdots + \beta_pX_{ip}$. In our experiments, the covariates are the features extracted from the reports.

$\psi$ can also be a non-linear risk function of a multilayer perceptron (MLP). To this end, our model is the same as DeepSurv \citep{Katzman_2018}.

Both CPH and DeepSurv cannot incorporate the higher-order temporal information from report sequences. To solve this problem, we define $\psi = LSTM(X_i)$ to model the possible time-variant effects of the covariates leading up to $t_u$ (Figure \ref{fig:architecture}). Our model is similar to RNN-SURV \citep{Giunchiglia2018RNNSURVAD} and DRSA \citep{ren2019deep}. The main difference is that the objective function is the average partial log-likelihood \cite{kvamme2019timetoevent}: 
\begin{equation}
-\frac{1}{N}{\sum_{i \in U_l}\left(\tilde{\psi}(x_i)-\log\sum_{j \in 
R_l}{e^{\tilde{\psi}(x_j)}}\right)}
\end{equation}

$U_l$ is  the set of patients that are deceased or last known to be alive (censored) by time point $t_l$. $R_l$ is the set of all live and uncensored patients before $t_l$. $N$ is the total number of deceased patients in the dataset.

\section{Experiments}

\subsection{Data}

The dataset \citep{pandey} is a collection of thoracoabdominal CT reports in English for heart failure patients from the New York-Presbyterian/Weill Cornell Medical Center who were admitted and discharged with billing codes ICD-9 Code 428 or ICD-10 Code I50 from January 2008 and July 2018 (Table~\ref{tab:dataset}). It was reviewed by the institutional board and de-identified. We use each patient's three most recent reports or zero vectors for any missing ones. Their time-to-event is calculated as the number of days between the most recent report date and death date if deceased or the last follow-up date if censored. We perform simple preprocessing steps to confirm each patient has at least one report and nonnegative time-to-event.

\begin{table}[!htpb]
\vspace{-0.5em}
    \centering
    \begin{tabular}{lr}
\toprule
Characteristics & $n$\\
\midrule
Number of Patients & 11,971\\
\hspace{1em}30 days mortality  & 1,209\\
\hspace{1em}365 days mortality & 2,062\\
\hspace{1em}Total mortality & 2,602 \\
Number of Reports & 39,752\\
\hspace{1em}Avg number of words in reports & 306\\
\bottomrule
    \end{tabular}
    \vspace{-.5em}
    \caption{Dataset Overview.}
    \label{tab:dataset}
\vspace{-1.7em}
\end{table}

\begin{table*}[h]
\centering
\begin{tabular}{p{14em}@{\hspace{1ex}}cccc}
\toprule
\textbf{Model} & \textbf{C-index} & \textbf{C-index@30} & \textbf{AUC@30} & \textbf{AUC@365}\\
\midrule
CPH + Feature Set 1 & 0.499 $\pm$ 0.025  & 0.504 $\pm$ 0.032 & 0.503 $\pm$ 0.037 & 0.501 $\pm$ 0.028 \\
CPH + Feature Set 2 & 0.621 $\pm$ 0.014  & 0.632 $\pm$ 0.033& 0.642 $\pm$ 0.030 & 0.642 $\pm$ 0.030 \\
CPH + Hidden Features & 0.674 $\pm$ 0.023 & 0.696 $\pm$ 0.022 & 0.710 $\pm$ 0.022 & 0.697 $\pm$ 0.026 \\
MLP + Feature Set 1 & 0.502 $\pm$ 0.023 & 0.509 $\pm$ 0.026 &0.509 $\pm$ 0.030 & 0.501 $\pm$ 0.032 \\
MLP + Feature Set 2 & 0.658 $\pm$ 0.010 & 0.671 $\pm$ 0.025 & 0.685 $\pm$ 0.023 & 0.683 $\pm$ 0.008 \\
MLP + Hidden Features & 0.704 $\pm$ 0.017 & 0.726 $\pm$ 0.020 & 0.744 $\pm$ 0.019 & 0.734 $\pm$ 0.018 \\
LSTM + Sequential HF & \bf{0.709 $\pm$ 0.022} & \bf{0.733 $\pm$ 0.031} & \bf{0.752 $\pm$ 0.033} & \textbf{0.742 $\pm$ 0.023} \\
\bottomrule
\end{tabular}
\vspace{-.5em}
\caption{\label{tab:overall}
Evaluation results. Feature Set 1 - \citep{pandey}, Feature Set 2 - \citep{irvin:2019}, Sequential HF - sequential hidden features
}
\end{table*}

\begin{table*}[ht!]
\centering
\begin{tabular}{p{14em}@{\hspace{1ex}}cccc}
\toprule
\textbf{Model} & \textbf{C-index} & \textbf{C-index@30} & \textbf{AUC@30} & \textbf{AUC@365}\\
\midrule
BERT-Base \cite{devlin-etal-2019-bert}  & 0.603 $\pm$ 0.115 & 0.611 $\pm$ 0.123& 0.618 $\pm$ 0.134 & 0.620 $\pm$ 0.136\\
BioBert \cite{biobert}& 0.701 $\pm$ 0.021  & 0.714 $\pm$ 0.027 & 0.734 $\pm$ 0.029 & 0.739 $\pm$ 0.028\\
ClinicalBert \cite{clinicalbert} & 0.692 $\pm$ 0.019 & 0.705 $\pm$ 0.023  & 0.723 $\pm$ 0.025 & 0.727 $\pm$ 0.029 \\
BlueBert \cite{bluebert} & \textbf{0.713 $\pm$ 0.019} & \textbf{0.735 $\pm$ 0.024} & \textbf{0.755 $\pm$ 0.024} & \textbf{0.756 $\pm$ 0.021} \\
CheXbert \cite{smit:2020} & 0.709 $\pm$ 0.022 & 0.733 $\pm$ 0.031 & 0.752 $\pm$ 0.033 & 0.742 $\pm$ 0.023 \\
\bottomrule
\end{tabular}
\vspace{-.5em}
\caption{\label{tab:bert}
Evaluation results of LSTM + sequential hidden features using different BERT models.}
\vspace{0em}
\end{table*}

\subsection{Metrics}

To assess the discriminative accuracies of our models, we use the C-index \citep{harrell} and time-dependent area-under-the-curve (AUC) \citep{time_auc_heagerty}, some of the most commonly used evaluation metrics in clinical research \citep{Kamarudin, pencina, uno_hajime}. Intuitively, the C-index measures the extent to which the model is able to assign logical risk scores. An individual with shorter time-to-event $T$ should have a higher risk score $R$ than the ones with longer time-to-event. Formally, it is defined as:
\begin{equation}
C = \frac{\sum\limits_{i, j} I(T_i > T_j)\cdot I(R_i < R_j)\cdot d_j}{\sum\limits_{i, j}I(T_i > T_j)\cdot d_j}
\end{equation}
\vspace{-5mm}
\begin{align*}
I(c)= \begin{cases} 1 \: \text{if}  \:  c  \:  \text{is true} \\  0 \: \text{else} \end{cases}
d_j = \begin{cases} 1 \: \text{if}  \:  T_j \: \text{exists} \\  0  \: \text{else} \end{cases}
\end{align*}
Both C-index and AUC assign a random model 0.5 and a perfect model 1. We measure all-time C-index, C-index at 30 days (C-index@30), and AUC at 30 days and 365 days (AUC@30 and AUC@365) to show the models’ performances dealing with different time-to-events\footnote{\url{https://github.com/sebp/scikit-survival}}.

\subsection{Training}

We perform a grid search to find the optimal hyperparameters based on the metrics and use them for all configurations. The learning rate is set to 0.0001 with an Adam optimizer. We iterate the training process for 100 epochs with batch size 256 and early stop if the validation loss does not decrease. The dropout rate is 0.6. We perform five-fold cross-validation to produce 95\% confidence intervals for each metric. The training, validation and test splits are 70\%, 10\%, 20\%, respectively. We use pycox and PyTorch to implement the framework\footnote{https://github.com/havakv/pycox}. The end-to-end training takes about 30 minutes with NVIDIA Tesla P100 16 GB GPU, mainly due to feature extraction.

\subsection{Results \& Discussions}
Table~\ref{tab:overall} shows our experimental results with variations in covariates and survival analysis models. 

Our LSTM model with hidden features (LSTM + Hidden Features) achieves the best results (0.709 in C-index), 3.5\% and 0.5\% improvements over CPH + Hidden Features and MLP + Hidden Features. In contrast to the MLP, its data included the reports from patients' prior visits with more textual and higher-order temporal information. Nonetheless, the improvements are stll marginal, suggesting that the evaluation of the effectiveness of LSTM in survival analysis in this context would require more empirical evidence, particularly with more longitudinal text data.

We observe that the hidden features provide at least 5\% improvements over the other feature sets with both CPH and MLP. This indicates that the hidden features capture textual information more thoroughly than the predefined features for survival analysis.
 
We find that Feature Set 2, obtained with CheXbert \citep{smit:2020}, performs significantly better ($>10\%$ C-index) than Feature Set 1, obtained with the CNN model \citep{pandey}. With both CPH and MLP, Feature Set 1 yields around 0.5 in C-index and AUC, whereas Feature Set 2 shows prognostic value in the 0.62-0.69 range. The difference of the feature sets directly results in the performance difference. While Feature Set 1 and Feature Set 2 have overlapping features (atelectasis, cardiomegaly, pleural effusion, and pneumonia), Feature Set 1 is not as discriminatory as Feature Set 2. This observation informs us that much important textual information with prognostic value is likely lost between the feature sets.

Finally, we compare our model on BERT-Base variants. BERT-Base, CheXbert and BlueBert used ``uncased'' text. BioBert and ClinicalBert used ``cased'' text. BioBert was pretrained on PubMed abstracts.  ClinicalBert was initialized with BioBert's weights and further trained on MIMIC-III clinical notes. BlueBert was pretrained on both datasets altogether. Table~\ref{tab:bert} shows that all BERT variants (except the original BERT) capture the textual information more comprehensively than the predefined features and yield significantly more accurate predictions. Further, the models with more pertinence to radiology reports perform incrementally better. BlueBert outperforms others and improves on CheXbert slightly. This observation is consistent with the findings in \cite{bluebert}. These results illustrate that using hidden layer representations in survival analysis can generalize across deep learning models based on their areas of focus. 

\subsection{Conclusion \& Future Work}
Incorporating the textual information of clinical texts in survival analysis is challenging because of their unstructured format. Automatic extraction tools have a small set of features selected by experts and fail to capture the information fully and precisely. We show a novel method of using hidden layer representations of clinical texts as covariates for proportional hazards models. When applied to predicting all-cause mortality of heart failure patients, the results indicate that hidden features encapsulate more comprehensive and effective textual information than predefined features. 

We plan to explore the use of the attention mechanism to the input sequence and test the generalizability of this method with more datasets. In addition, we plan to gain more insights on how hidden features are influenced (e.g. word choice, text length, etc.) and add value for better prediction as interpretability is highly important in the medical domain. We hope our small contribution provides assistance in the scalable development of accurate predictive models that harness clinical text information. 

\section*{Acknowledgment}

Research reported in this publication was supported by National Library of Medicine - National Institutes of Health (NIH) under award number R00LM013001. It was also supported by National Heart, Lung, and Blood Institute of NIH under award number R01HL1276610. This study also received support from NewYork-Presbyterian Hospital (NYPH) and Weill Cornell Medical College (WCMC), including the Clinical and Translational Science Center (CTSC) (UL1TR002384) and Joint Clinical Trials Office (JCTO). Additionally, we would like to thank Dr. Pranav Rajpurkar and Dr. Matthew P Lungren for providing the ChexBert model.

\bibliography{anthology,custom}
\bibliographystyle{acl_natbib}

\end{document}